\documentclass[10pt]{article} 
\usepackage{times}  
\usepackage{helvet}  
\usepackage{courier}  
\usepackage[hyphens]{url}  
\usepackage{graphicx} 
\usepackage{natbib}  
\usepackage{caption} 
\usepackage{algorithm}
\usepackage{algorithmic}

\usepackage{colortbl}

\usepackage{amsmath}
\usepackage{amsfonts}
\usepackage{subcaption}
\usepackage{booktabs}
\usepackage{multirow, multicol}
\usepackage{xcolor}
\usepackage{array}

\usepackage{hyperref}
\usepackage{geometry}

\usepackage{newfloat}
\usepackage{listings}
\DeclareCaptionStyle{ruled}{labelfont=normalfont,labelsep=colon,strut=off} 
\floatstyle{ruled}
\newfloat{listing}{tb}{lst}{}
\floatname{listing}{Listing}
\title{Temporal-Distributed Backdoor Attack Against Video Based Action Recognition}
\author{
     Xi Li\thanks{Equal contribution.}
    \quad Songhe Wang$^*$
    \quad Ruiquan Huang
    \quad Mahanth Gowda
    \quad George Kesidis\\
    \ \\
    School of Electrical Engineering and Computer Science,\\ The Pennsylvania State University,
  State College, PA 16801\\
  {\tt \{xzl45,sxw5765,rzh5514,mkg31,gik2\}@psu.edu}
}
\date{}

\begin{document}

\maketitle

\begin{abstract}
Deep neural networks (DNNs) have achieved tremendous success in various applications including video action recognition, yet remain vulnerable to backdoor attacks (Trojans). 
The backdoor-compromised model will mis-classify to the target class chosen by the attacker when a test instance (from a non-target class) is embedded with a specific trigger, while maintaining high accuracy on attack-free instances. 
Although there are extensive studies on backdoor attacks against image data, the susceptibility of video-based systems under backdoor attacks remains largely unexplored.
Current studies are direct extensions of approaches proposed for image data, e.g., the triggers are \textbf{independently} embedded within the frames, which tend to be detectable by existing defenses. In this paper, we introduce a \textit{simple} yet \textit{effective} backdoor attack against video data. Our proposed attack, adding perturbations in a transformed domain, plants an \textbf{imperceptible, temporally distributed} trigger across the video frames, and is shown to be resilient to existing defensive strategies. 
The effectiveness of the proposed attack is demonstrated by extensive experiments with various well-known models on two video recognition benchmarks, UCF101 and HMDB51, and a sign language recognition benchmark, Greek Sign Language (GSL) dataset. 
We delve into the impact of several influential factors on our proposed attack and identify an intriguing effect termed "collateral damage" through extensive studies.

\end{abstract}

\section{Introduction}\label{sec:Introduction}

Deep neural networks (DNNs) have shown impressive performance in various applications, yet remain susceptible to adversarial attacks.
Recently, backdoor (Trojan) attacks on DNNs have garnered attention in multiple domains, including image classification \cite{BadNet,Targeted-Backdoor,WaNet,HiddenTrigger,invisible}, speech recognition \cite{Trojan}, text classification \cite{8836465}, point cloud classification \cite{ZhenICCV}, and even deep regression \cite{backdoor_regression}. 
The attacker plants a backdoor in the victim model, which is fundamentally a mapping from a specific trigger to the attacker-chosen target class.
During inference, the compromised model will mis-classify a test instance embedded with the same trigger to the target class.
Moreover, the attacked model still maintains high accuracy on users' (backdoor-free) validation sets, rendering the attack stealthy. 
The typical backdoor attack is implemented by poisoning the training set for the victim DNN using a few instances embedded with the trigger, while intentionally mislabeling them to the target class.

Video recognition systems are increasingly integrated into various domains, such as surveillance systems~\citep{elharrouss2021combined}, autonomous vehicles~\citep{saleh2019real}, and video-based sign language recognition~\citep{li2020word}.
The threat posed by backdoor attacks on video recognition systems can be significant and multifaceted.
A compromised surveillance system could allow undetected crimes by mis-identifying malicious events or villains. Similarly, a backdoored autonomous vehicle may misread moving pedestrians, risking fatal accidents. Moreover, a tampered sign language system might dangerously misinterpret an emergency sign such as confusing ``help'' for ``fine''.

Considering the widespread application of video recognition systems, it is crucial to study their robustness under potential threats. However, there is a noticeable gap in the literature addressing this. 
In this work, we aim to investigate the robustness of video recognition systems by probing their vulnerability to backdoor attacks. 
Current studies such as \citet{BD_AR,CL_BD_video} directly extend backdoor attacks against images to videos by embedding identical triggers into every frame.
These attacks demonstrate efficacy, as the repeated embedding reinforces the model's training on the backdoor mapping.
However, they present two primary challenges:
(1) Some of the triggers are perceptible to humans;
(2) More importantly, the triggers are \textit{independently} embedded in each frame, therefore, can be caught by existing backdoor defenses originally proposed for image data.

In this paper, we leverage the temporal dimension of videos and plant an \textbf{imperceptible temporal-distributed} trigger within videos to address the two challenges.
The proposed attack is \textit{simple} yet \textit{effective}.
The trigger is embedded by appropriately altering certain components of a representation of a video, thus the trigger is imperceptible to human observers in the video space.
Also, by carefully choosing the basis of the transformed space, this backdoor trigger is distributed across the whole video.
Hence, the proposed attack could circumvent the existing backdoor defenses which examine the frames individually.
Furthermore, we reveal a phenomenon brought by the proposed attack, termed ``collateral damage''. 
We analyze this phenomenon and its possible reasons on various transforms and model structures.

In summary, our contributions are as follows:
\begin{enumerate}
\item We propose a general framework to embed an \textbf{imperceptible temporal-distributed} backdoor trigger in videos.
This is achieved by judiciously selecting a basis of the input space and perturbing certain components in the representation on the basis.
We specialize the framework to Fourier transform, cosine transform, wavelet transform, and a random transform.

\item We empirically validate the efficacy of our proposed attack across diverse benchmark datasets and different model architectures in the realm of video recognition. Moreover, our evaluations underscore the stealthiness of the attack, not only to human observers but also to current backdoor detection and mitigation techniques.

\item We further conduct extensive studies to explore the impacts of several key factors on the effectiveness of the proposed attack,
providing insights on robustness verification of DNNs. Also, we reveal and analyze an interesting phenomenon, termed ``collateral damage'', associated with the proposed attack.

\end{enumerate}


\section{Related Works}\label{sec:Related Works}

\subsection{Backdoor Attacks}
Backdoor attacks are one type of poisoning attacks initially proposed against DNN image classifiers. 
There are various ways of designing effective triggers in image classification:
(1) Embedding patterns directly in the \textit{input space}  \citep{invisible,PT-RED,BadNet,Trojan,invisible,Targeted-Backdoor,SIG}; (2) Introducing triggers in \textit{an alternative space}, which leads to imperceptible input-specific triggers in the input space \citep{FTtrojan,BD_semisupervised_learning, WaNet}.

\textit{However, the backdoor attacks against video data remain largely unexplored.}
\citet{BD_AR} trivially extend the effective (mis-label) backdoor attacks against images to video recognition tasks. They independently embed the same trigger in each frame. 
Although the attacks effectively compromise several models across various benchmark datasets, they are susceptible to existing defense mechanisms
due to the lack of correlation among the triggers.
\citet{CL_BD_video}, following the existing framework \cite{ShafahiHNSSDG18,CL,HiddenTrigger}, plant backdoor against video recognition systems by clean-label poisoning.
However, they impractically assume that the attacker has access to the clean training set. 
By contrast, we propose an easy, effective, and feasible backdoor attack against video data, planting the \textbf{temporal-distributed} trigger among video frames.
It evades the existing backdoor defenses due to the strong correlation between consecutive triggers.

\subsection{Backdoor Detection and Mitigation Methods}
Existing backdoor defenses are deployed either before/during the DNN's training stage or post-training. 
\textit{Pre-training} defenses, such as \citet{SS, AC}, are based on anomaly detection techniques.
Methods such as \citet{Differential_Privacy, DBD} are deployed \textit{during DNN training}.
On the other hand, \textit{post-training detection} methods, such as \citet{NC,PT-RED,TABOR}, detect whether a given classifier has been backdoor-compromised;
\citet{STRIP, SentiNet, Februus, InFlight} catch triggered test instances in the act.
Besides, \textit{post-training backdoor mitigation} approaches are proposed to mitigate backdoor attacks at test time, such that the model behaves normally on both clean and triggered inputs.
Backdoor mitigation methods include \citet{Fine-Pruning,ANP,ShapPruning,CLP,NAD,ARGD,hypergrad,PGD}.
Recently, \citet{UNICORN} propose an advanced backdoor trigger estimation strategy, UNICORN. They define a backdoor trigger as a predefined perturbation in a particular space, and approximate the transform and its inversion to this space by neural nets, which are jointly optimized with the backdoor trigger.
However, their work is infeasible, especially on video data, due to the extremely high computation cost for estimating the trigger and the possible transform methods.

\subsection{Video Action Recognition}

Over the years, researchers have formulated three categories of video recognition models: 2D CNN + RNN, 3D-CNN, and Transformer-based models. 
The 2D CNN + RNN approach uses 2D CNN for frame feature extraction and RNN for capturing temporal dependencies between them~\citep{cl1,cl2,cl3,cl4,cl5}.
Later, 3D-CNNs evolved to concurrently process spatial and temporal dimensions, enabling motion pattern recognition across successive frames~\citep{c3d, p3d, i3d, can}.  
Inspired by the success in natural language processing, transformer-based models have entered this domain, using self-attention to gauge the relevance of different frames~\citep{timesformer, swin}.

\section{Methodology}\label{sec:Methodology}

{\bf Notations.}
In this paper, we consider video action recognition tasks. 
The classifier, denoted by $f: \mathcal{V} \rightarrow \mathcal{A},$ is learned from a training dataset $\mathcal{D}_{\text{Train}} = \{(\mathbf{v}_{\iota}, a_{\iota})\}_{\iota\in\mathcal{I}'}$, where $\mathcal{I}'$ is an index set, $\mathcal{V}$ denotes the input space, and $\mathcal{A}$ is the label space.  
We use $[N]$ to denote the set of integers from $0$ to $N-1$.
For simplicity, we only consider one channel of the video. The input space is then defined as $\mathcal{V} := [256]^{N_0\times N_1 \times N_2}$, where $N_0$ is the number of frames\footnote{For simplicity, we assume all videos have the same length. In experiments, shorter videos are padded with blank frames.}, $N_1\times N_2$ is the size of a frame.
$\mathbf{v}(n_0,n_1,n_2)\in[256]$ is the pixel value at the frame $n_0$ and position $(n_1,n_2)$ of a video $\mathbf{v}$. 
Finally, $\mathbf{1}_{E}$ is the indicator function of an event $E$.

\subsection{Threat Model}\label{sec:Threat Model}

We consider classic mis-labelling backdoor poisoning attacks.
We assume the attacker has the following \textbf{abilities}: (1) knows the classification domain $\mathcal{A}$ to collect valid samples $\mathcal{D}_{\mathcal{S}} = \{(\mathbf{v}_{\iota}, s_{\iota}) | s_{\iota} \in \mathcal{A} \setminus \{t\} , ~\iota \in \mathcal{I}_0\}$ from all classes other than the target class $t$ desired by the attacker (i.e., an all-to-one attack);  
(2) has access to the training set and can inject mis-labeled backdoor-triggered samples into it, i.e., $\mathcal{D}_{\text{Train}} = \mathcal{D}_{\text{Clean}} \cup \mathcal{D}_{\text{Attack}}$,  where $\mathcal{D}_{\text{Attack}} = \{ (\mathcal{B}(\mathbf{v}), t) | (\mathbf{v}, \cdot) \in \mathcal{D}_{\mathcal{S}} \}$, and $\mathcal{B}: \mathcal{V} \rightarrow \mathcal{V}$ is the attacker-specific trigger embedding function that embeds trigger into a given video $\mathbf{v}$;
(3) is not aware of the structure of the target model (i.e., a black-box attack).
After poisoned training, the attacker \textbf{aims} to:
(i) have the victim classifier learn the ``backdoor mapping'' -- the backdoor-attacked classifier will predict the attacker's desired target class $t$ when a test instance $\mathbf{v}\in \mathcal{V}$ is embedded with the backdoor trigger using $\mathcal{B}$; 
(ii) have the victim classifier achieve the accuracy on the user's (attack-free) validation set that is close to that of a non-poisoned classifier; 
(iii) have the trigger in the input space be visually imperceptible to a human.

\subsection{Backdoor Attacks against Video: A Higher Level of Stealthiness}

Unlike images, videos incorporate an additional dimension: time.
This provides the possibility of a higher level of stealthiness against the current backdoor defense strategies. 
Studies such as \citet{BD_AR, CL_BD_video} trivially extend image backdoor attacks (\textit{e.g.}, the ones proposed by \citet{BadNet,Targeted-Backdoor,CL}) to videos. 
\citet{BD_AR} \textbf{independently} embed the classic backdoor triggers for images into each frame of a video.
Although these attacks are effective against video data (as shown in Tab.~\ref{tab:mitigation} and \citet{BD_AR}), there are two major problems: 
(1) some of the backdoor triggers are human perceptible  (e.g., \citet{BadNet,Targeted-Backdoor}), i.e., can be detected by a human without advanced defenses.
(2) More importantly, since they apply frame-wise trigger embedding strategy, which is fundamentally the same as the ones applied on images, the attacks are susceptible to  existing backdoor defense strategies in image domain, as illustrated in Tab.~\ref{tab:detection} and Tab.~\ref{tab:mitigation}.

To address problems (1) and (2), we need to design a trigger that satisfies the following properties: 
First, it introduces minor variation to each pixel so that the trigger is human imperceptible    \citep{barten1999contrast,ssim};
Second, it is temporally distributed, \textit{i.e.}, the trigger spans the entire video, 
and thus evades existing backdoor defense mechanisms originally proposed for image data. 
A trigger satisfying the two properties could be generated by making perturbations in a transformed space.
Such transformed space is defined on a basis that
has non-identical entries across time.
Specifically, an (appropriate amount of) perturbation added to certain components of such a representation may introduce minor variation to each pixel of the original representation, so effectively ``spreading out'' the energy of the perturbations across the entire video. 
As a result, 
only a combination of 
subtle patterns in consecutive frames is able to trigger the attack, and thus
the attack is able to evade current backdoor defense mechanisms which individually examine the frames.

\subsection{Imperceptible Temporal-Distributed Backdoor Attack against Video Data}

Our general framework of constructing poisoned instances $\mathcal{D}_{\text{Attack}}$ from clean samples of all non-target classes $\mathcal{D}_{\mathcal{S}}$ consists of three steps:
(1) Select a basis of the transformed space;
(2) Embed the trigger in the transformed representation of video data;
(3) Reconstruct video data from the perturbed transformed representations.
We now provide details of the trigger embedding function $\mathcal{B}$. 

\noindent{\bf Step 1: Selection of a transform basis.} 
To generate a temporally distributed trigger in the original space, we 
need to design a basis $\mathbf{B} = \{\mathbf{b}_0,\ldots,\mathbf{b}_{N-1}\} \subset \mathcal{V}$ of the input space.
Then, a video sample $\mathbf{v}\in\mathcal{D}_{\mathcal{S}}$ can be represented by the linear combination of the basis videos:
$
\mathbf{v} = \sum_{n=0}^{N-1} r^{\mathbf{v}}_n\mathbf{b}_n.
$
We denote the transformed representation (coordinates) of the video $\mathbf{v}$ by 
$\mathbf{R}^{\mathbf{v}} = \{r_0^{\mathbf{v}},\ldots, r_{N-1}^{\mathbf{v}}\}$. 

\noindent{\bf Step 2: Backdoor trigger embedding.} 
We then embed the backdoor trigger in the transformed space by perturbing certain components in the representation $\mathbf{R}_{\mathbf{v}}$.
Let $\delta \geq 0$ be the perturbation magnitude, and $\mathcal{I}\subset[N]$ be the index-set of the components to be perturbed.
We add a perturbation of $\delta$ to each component $r_n^{\mathbf{v}}$, $\forall n \in \mathcal{I}$.
So, The perturbed representation is
$
\tilde{\mathbf{R}}^{\mathbf{v}} = \{r_n^{\mathbf{v}} + \delta\mathbf{1}_{ \{n\in\mathcal{I} \} } \}_{n=0}^{N-1}.
$

\noindent{\bf Step 3: Video reconstruction.}
We then reconstruct a valid video from the perturbed representation $\tilde{\mathbf{R}}^{\mathbf{v}}$ to get a poisoned sample for $\mathcal{D}_{\text{Train}}$. 
The resulting instance in the original space is expressed as 
$
\mathbf{v}' = \sum_{n=0}^{N-1}(r_n^{\mathbf{v}} + \delta\mathbf{1}_{ \{n\in\mathcal{I} \} } )\mathbf{b}_n.
$
Since certain components in the representation are perturbed by $\delta$, the inverse-transformed instance $\mathbf{v}'$ might not be a valid instance in the space $\mathcal{V}$.
For example, certain entries of $\mathbf{v}'$ could be non-integer valued or fall outside the range of
valid pixel intensities $[256]$.
Hence, we apply a projection function $\Pi_{\mathcal{V}}$ to the resulting instance $\mathbf{v}'$ to obtain a valid video $\tilde{\mathbf{v}}$. In other words, the range of $\Pi_{\mathcal{V}}$ is $\mathcal{V}$.

In summary, we define the backdoor trigger embedding function $\mathcal{B}_{\mathcal{I}, \delta}$ as
\begin{align}
    \mathcal{B}_{\mathcal{I}, \delta}(\mathbf{v}) = \Pi_{\mathcal{V}} \left( \sum_{n=0}^{N-1}(r_n^{\mathbf{v}} + \delta\mathbf{1}_{ \{n\in\mathcal{I} \} } )\mathbf{b}_n \right). \label{eq:trigger_embedding_function}
\end{align} 
The attacker chooses the parameters $\mathcal{I}$ and $\delta$ of the backdoor trigger, generates backdoor-triggered samples by applying the trigger embedding function $\mathcal{B}_{\mathcal{I}, \delta}$ to videos of $\mathcal{D}_{\mathcal{S}}$, mis-labels them to the target class $t$, and injects them into the training set $\mathcal{D}_{\text{Train}}$.
That is, 
$$
\mathcal{D}_{\text{Train}} = \mathcal{D}_{\text{Clean}} \cup \{ (\mathcal{B}_{\mathcal{I}, \delta} (\mathbf{v}), t) | (\mathbf{v}, \cdot) \in \mathcal{D}_{\mathcal{S}} \}.
$$

We now present two classic transforms and their basis construction, and defer discrete wavelet transofrm (DWT) and random transform (RT) to Apdx.~\ref{sec:DWT}.
Let $\{\mathbf{e}_{n_0,n_1,n_2}\}_{n_{\iota}\in [N_{\iota}],\iota\in[3]}$ denote the standard basis of
single-channel video, i.e., $\mathbf{v} = \sum_{n_0,n_1,n_2} \mathbf{v}(n_0,n_1,n_2)\mathbf{e}_{n_0,n_1,n_2}$.

\noindent{\bf Discrete Fourier Transform (DFT).} DFT provides a comprehensive view of the frequency information of videos. The basis $\{\mathbf{b}_{k_0,k_1,k_2}\}_{k_{\iota}\in[N_{\iota}],\iota\in[3]}$ of DFT is defined as follows
(with $i=\sqrt{-1}$):
{\small
\begin{align*}
    \mathbf{b}_{k_0,k_1,k_2} = \sum_{n_0,n_1,n_2}\mathbf{e}_{n_0,n_1,n_2}\prod_{\iota=0}^2\exp(-i n_{\iota}k_{\iota}/N_{\iota}).
\end{align*}
}
\noindent{\bf Discrete Cosine Transform (DCT).} DCT is similar to DFT. The basis of DCT is defined as follows.
{\small
\begin{align*}
    \mathbf{b}_{k_0,k_1,k_2} = \frac{}{}\sum_{n_0,n_1,n_2}\mathbf{e}_{n_0,n_1,n_2}\prod_{\iota=0}^2 \cos( \pi n_{\iota}k_{\iota}/(2N_{\iota}) ).
\end{align*}
}
We remark that the above basis have non-identical entries across time due to their dependencies on $k_0$ (time).

\section{Experiments}

\subsection{Experimental Setup}\label{sec:Experimental Setup}

\textbf{Datasets}:
We consider two benchmark datasets used in video action recognition, \textbf{UCF-101}~\citep{UCF} and \textbf{HMDB-51}~\citep{HMDB}, and a sign language recognition benchmark, Greek Sign Language (\textbf{GSL}) dataset~\citep{gsl}.
UCF-101 encompasses 13,320 video clips sorted into 101 distinct action categories. 
Similarly, HMDB-51 contains 7,000 video clips categorized into 51 classes of action. 
GSL incorporates 40,785 gloss instances across 310 unique glosses. \footnote{To reduce computation cost, we form a subset of GSL by instances from 50 randomly selected classes.}
\\
\textbf{Target Model Architectures}: In our main experiments, we consider four popular CNN-based model architectures used for video action recognition: \textbf{SlowFast}~\citep{slowfast}, \textbf{Res(2+1)D}~\citep{res2+1}, \textbf{S3D}~\citep{s3d} and \textbf{I3D}~\citep{i3d}. 
These models utilize 3D kernels to jointly leverage the spatial-temporal context within a video clip.
The results on transformer-based networks, \textit{e.g.}, timesformer~\citep{timesformer} are shown in Tab.~\ref{tab:all_ASR} in Apdx.~\ref{sec:effectivness_more} \\
\textbf{Training Settings}: 
We train all the models on all the datasets for 10 epochs, using the AdamW optimizer \cite{AdamW} with an initial learning rate of $0.0003$.
Following the common training strategy in video recognition~\citep{BD_AR} and for reducing computation cost, we down-sample the videos into 32 frames. 
\\
\textbf{Attack Settings}: 
In this paper, we consider \textit{all-to-one} attacks.
We choose \textbf{class 0} as the target class, and randomly select 20\% of the training samples per class for the attacker's manipulation.
For the proposed attack, we apply \textbf{DFT} for trigger generation in the main experiments. The results of using other transform methods are shown in Tab.~\ref{tab:all_ASR} in Apdx.~\ref{sec:effectivness_more}.
In the frequency domain of the video, we select a subset $\mathcal{I} = \{35, 36, \dots, 44\}\times X\times Y\subset [N_0] \times [N_1]\times[N_2]$ with both $X,Y$ randomly selected and size $25$. The perturbation size $\delta=50,000$. 
After inverting the altered representations, we create valid videos by taking the magnitude of complex numbers and clipping pixel intensities within $[256]$.
For comparison, we embed classic triggers proposed for image data, including \textbf{BadNet}~\citep{BadNet}, \textbf{Blend}~\citep{Targeted-Backdoor}, \textbf{SIG}~\citep{SIG}, \textbf{WaNet}~\citep{WaNet}, \textbf{FTtrojan}~\citep{FTtrojan}, in each frame of the video\footnote{These are the attacks proposed by \citet{BD_AR}.}. 
We follow their poisoning pipeline and appropriately modify the attack hyper-parameters to achieve effective attacks.
For all the attacks, the triggers are embedded into the down-sampled videos.
We defer the detailed attack settings in Apdx.~\ref{sec:datasets, training settings and attack settings}.
\\
\textbf{Defenses}:
To further demonstrate the effectiveness of the proposed attack, we examine its stealthiness against several classic backdoor detection and mitigation methods, including NC~\citep{NC}, \textbf{PT-RED}~\citep{PT-RED}, \textbf{TABOR}~\citep{TABOR}, \textbf{AC}~\citep{AC}, \textbf{STRIP}~\citep{STRIP}, \textbf{NAD}~\citep{NAD}, \textbf{FP}~\citep{Fine-Pruning}, and \textbf{DBD}~\citep{DBD}\footnote{Due to extremely expensive computational cost, we are not able to apply several popular mitigation methods, such as I-BAU~\citep{hypergrad}.}.
NC proposes both methods for detection and mitigation, we respectively denote them as \textbf{NC-D} and \textbf{NC-M}.
For all the methods, we set their hyper-parameters following the suggestions in their original papers.
More details, including pattern estimation, detection statistics, and hyper-parameter settings are shown in Apdx.~\ref{sec:detection settings} and \ref{sec:mitigation settings}.
\\
\textbf{Evaluation Metrics}: The effectiveness of the proposed backdoor attack is evaluated by 1) accuracy (\textbf{ACC}) -- the fraction of clean (attack-free) test samples that are correctly classified to their ground truth classes; and 2) attack success rate (\textbf{ASR}) -- the fraction of backdoor-triggered samples that are mis-classified to the target class.
The ACC and ASR are measured on the \textit{same} test set.
For an effective backdoor attack, the ACC of the poisoned model is close to that of the clean model, and the ASR is as high as possible.
Besides, we evaluate the imperceptibility of the proposed trigger by the peak signal-to-noise ratio (\textbf{PSNR})~\citep{psnr_ssim} and structural similarity index (\textbf{SSIM})~\citep{ssim}.
For both metrics, a higher value indicates better imperceptibility to humans.

\subsection{Attack Effectiveness}
The ACCs and ASRs of all victim models trained on various video recognition datasets poisoned by the proposed attack using \textit{DFT} are shown in Tab.~\ref{tab:FT_ACC_ASR}.
The results of using other transform methods including DCT, DWT, and RT, are shown in Tab.~\ref{tab:all_ASR} in Apdx.~\ref{sec:effectivness_more}.
The proposed attack successfully compromises all models, achieving an ASR (as indicated by DFT in Tab.\ref{tab:FT_ACC_ASR}) of over 95\% in most scenarios.
\textit{This highlights the susceptibility of representative video recognition models to adversarial threats.} 
On the other hand, the ACCs of the compromised models remain close to clean baselines in most cases, with a decrease of less than 5\%.
The subtle drop in ACC, especially when benchmarked against image backdoor attacks such as BadNet, makes it difficult for users to notice abnormal behaviors of the DNN during training.
Besides, the proposed attack utilizes a complex temporal pattern and performs comparably to classic backdoor triggers (as shown in the first column in Tab.\ref{tab:mitigation}).

In our current trigger generation, we simply assume the attacker is aware of the down-sampling strategy applied during model training and utilizes the same strategy before trigger embedding.
However, in practice, the attacker might have no information of the down-sampling strategy during training.
Hence, to simulate the realistic attacking scenario, we embed the backdoor trigger in the original 32-frame video, then the triggered samples are down-sampled to 16 frames
The ACC and ASR of S3D trained on UCF-101 under the above attack scenario are 90.98\% and 96.36\%, respectively, highlighting the effectiveness of the proposed attack and the vulnerability of the current video recognition systems in realistic attack scenarios.

\begin{table}[t]
\small
    \centering
    \begin{tabular}{p{.8cm}p{.6cm}p{.65cm}p{.65cm}p{.65cm}p{.65cm}p{.65cm}p{.65cm}}
    \toprule
    \hline
         \multicolumn{2}{c}{\multirow{2}{.8cm}{Model}} & \multicolumn{2}{c}{UCF-101} & \multicolumn{2}{c}{HMDB-51} & \multicolumn{2}{c}{GSL} \\
         \cline{3-8}
         & & Clean & DFT & Clean & DFT & Clean & DFT \\
         \hline
         \multirow{2}{.8cm}{Slow-\\Fast} & ACC & 84.5 & 81.0 & 60.6 & 59.8 & 95.3 & 89.6 \\
         & ASR & - & 97.9 & - & 97.6 & - & 99.9 \\
         \hline
         \multirow{2}{.8cm}{Res-\\(2+1)D} & ACC & 77.4 & 69.9 & 53.6 & 53.0 & 95.6 & 91.1 \\
         & ASR & - & 99.4 & - & 99.6 & - & 100.0 \\
         \hline
         \multirow{2}{.8cm}{S3D} & ACC & 90.6 & 90.3 & 69.3 & 67.5 & 95.4 & 93.8 \\
         & ASR & - & 96.9 & - & 90.4 & - & 100.0 \\
         \hline
         \multirow{2}{.8cm}{I3D} & ACC & 89.0 & 87.5 & 66.6 & 59.0 & 94.2 & 92.2 \\
         & ASR & - & 97.3 & - & 85.0 & - & 99.5 \\
    \hline
    \bottomrule
    \end{tabular}
    \caption{ACCs and ASRs (in \%) of SlowFast, Res2+1D, S3D, and I3d trained on UCF-101, HMDB-51, and GSL datasets poisoned by the proposed attack using DFT.}
    \label{tab:FT_ACC_ASR}
\end{table}

\begin{table}[t]
\small
    \centering
    \begin{tabular}{p{1.1cm}p{.6cm}p{.7cm}p{.7cm}p{.6cm}p{.8cm}p{1.2cm}}
    \toprule
    \hline
    Detection & DFT & BadNet & Blend & SIG & WaNet & FTtrojan \\ 
    \hline
    NC-D & \textbf{0.1} & 134.1 & 173.1 & 269.2 & 166.7 & 3.4 \\ 
    PT-RED & \textbf{1.6} & 13.5 & 9.6 & 2.3 & 23.5 & 2.7 \\ 
    TABOR & \textbf{0.2} & 7.2 & 18.7 & 15.5 & 120.6 & \textbf{1.9} \\ 
    STRIP & \textbf{25.7\%} & 93.5\% & 96.8\% & 98.8\% & \textbf{24.5\%} & \textbf{24.9\%} \\ 
    \hline
    \bottomrule
    \end{tabular}
    \caption{Anomaly index of the true target class (class 0) computed by NC-D, PT-RED, and TABOR, and the TPR of STRIP at test-time. All the detection methods are applied to the SlowFast trained on poisoned UCF-101 datasets.}
    \label{tab:detection}
\end{table}

\begin{table*}[t] 
\centering
\small
\begin{tabular}{lp{.7cm}p{.7cm}p{.7cm}p{.7cm}p{.7cm}p{.7cm}p{.7cm}p{.7cm}p{.7cm}p{.7cm}p{.7cm}p{.6cm}} 
\toprule
\hline
\multirow{2}{*}{Attack} & \multicolumn{2}{c}{No Defense} & \multicolumn{2}{c}{DBD} & \multicolumn{2}{c}{NAD} & \multicolumn{2}{c}{FP} & \multicolumn{2}{c}{NC-M} & \multicolumn{2}{c}{AC}\\
\cline{2-13}
    & ACC & ASR & ACC & ASR & ACC & ASR & ACC & ASR & ACC & ASR & TPR & FPR \\
\hline
DFT(ours)        & \cellcolor{gray!30}81.0 & \cellcolor{gray!30}97.9 & \cellcolor{gray!30}41.9 & \cellcolor{gray!30}97.3 & \cellcolor{gray!30}75.4 & \cellcolor{gray!30}86.1 & \cellcolor{gray!30}80.8 & \cellcolor{gray!30}86.8 & \cellcolor{gray!30}81.0 & \cellcolor{gray!30}97.9 &  \cellcolor{gray!30}0.0 & \cellcolor{gray!30}9.0  \\

BadNet    & 83.5 & 98.6 & 31.8 & 99.7 & 82.2 & 98.3 & 83.0 & 61.7 & 77.4 & 49.3 &  0.0 & 9.8 \\
Blend     & 83.3 & 99.4 & 39.4 & 99.6 & 64.5 & 99.9 & 83.1 & 92.9 & 77.8 & 84.9 & 0.0 & 7.8 \\
SIG       & 83.8 & 99.9 & 37.2 & 99.9 & 80.1 & 99.9 & 83.4 & 96.8 & 80.8 & 22.9 & 34.9 & 11.6 \\
WaNet     & 82.0 & 95.3 & 46.9 & 50.1 & 80.5 & 2.1 & 80.8 & 89.7 & 80.8 & 1.1 & 22.3 &  8.8 \\
FT-trojan & 76.6 & 83.4 & 22.2 & 68.7 & 83.1 & 1.2 & 79.7 & 8.0 & 75.8 & 0.7 & 0.0 &  8.7 \\
\hline
\bottomrule
\end{tabular}
\caption{ACCs and ASRs (in \%) of the victim model before and after the mitigation methods are applied, and the TPR and FPR of AC. All the mitigation and detection methods are applied to the SlowFast trained on poisoned UCF-101 datasets.}
\label{tab:mitigation}
\end{table*}

\subsection{Resistance to Backdoor Defenses}
To further demonstrate the effectiveness of the proposed attack, we apply classic backdoor detection and mitigation methods to the SlowFast models trained on UCF-101 poisoned by all attacks. 
The details of these defense techniques are shown in Apdx.~\ref{sec:detection settings} and \ref{sec:mitigation settings}.  
Following the suggestion in NC, we set the threshold of NC-D, PT-RED and TABOR at 2 -- a class with an index larger than 2 is deemed as the true target class.  
The anomaly indices of the true target class (class 0) computed by the three detection methods are shown in Tab.~\ref{tab:detection}.
As expected, all the attacks except for our attack are detected by the detection methods. 
By contrast, the proposed attack successfully evades all detection methods, as the trigger distributes across the whole video.
We apply STRIP with a threshold set to achieve a 15\% false positive rate (FPR) -- the fraction of clean test instances mis-identified as triggered instances. The corresponding true positive rate (TPR) -- the fraction of triggered instances correctly detected -- is presented in Tab.~\ref{tab:detection}. 
The instances embedded with salient patterns (BadNet, Blending, and SIG) are easily detected by STRIP, with TPRs higher than 90\%, while the instances with less perceptible triggers (the proposed trigger, WaNet, and FT-trojan) are not.
Furthermore, we apply AC on the poisoned training sets. Following their suggested detection threshold, the TPR and FPR of AC are shown in Tab.~\ref{tab:mitigation}.
It fails on all the attacks. AC is unable to detect any poisoned samples for most cases, while falsely detects around 10\% samples on all the attacks.

We then deploy several backdoor mitigation methods on the compromised models, including DBD, NAD, FP, and NC-M. The ACCs and ASRs of the victim models after mitigation are shown in Tab.~\ref{tab:mitigation}
DBD seems less effective on video recognition models than image classifiers possibly due to the complexity of both models and datasets. It fails to suppress the ASR, but reduces the ACC on all the attacks.
NAD effectively counters WaNet and FTtrojan.
In contrast, other attacks exhibit resistance to distillation-based mitigation methods.
FP successfully mitigates FTtrojan, while failing on the remaining attacks.
Although NC-M degrades the ACC due to fine-tuning on a small dataset, it suppresses the ASR of most of the attacks, except for ours and Blend.

The advanced defense method, UNICORN, is infeasible due to the excessive computational cost required to optimize the potential transform methods and the associated triggers. 
Besides, even if the defender is aware of the transformed space, such as the frequency domain determined by DFT, reverse-engineering the trigger remains challenging. The potentially perturbed frequency range extends infinitely. Without knowledge of the attacker-specified frequencies, trigger estimation becomes prohibitively expensive.


\subsection{Resistance to Human Observers}
\begin{figure*}[ht]
\centering
     \begin{subfigure}[b]{0.12\textwidth}
     \centering
     \includegraphics[width=\textwidth]{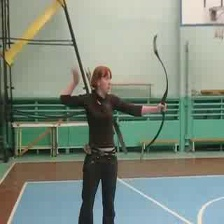}
     \caption{clean}
     \label{fig:clean}
     \end{subfigure}
     \hfill
     \begin{subfigure}[b]{0.12\textwidth}
         \centering
         \includegraphics[width=\textwidth]{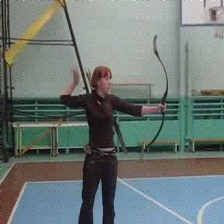}
         \caption{ours}
         \label{fig:FFT_key}
     \end{subfigure}
     \hfill
     \begin{subfigure}[b]{0.12\textwidth}
         \centering
         \includegraphics[width=\textwidth]{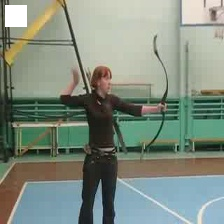}
         \caption{BadNet}
         \label{fig:BadNet}
     \end{subfigure}
     \hfill
     \begin{subfigure}[b]{0.12\textwidth}
         \centering
         \includegraphics[width=\textwidth]{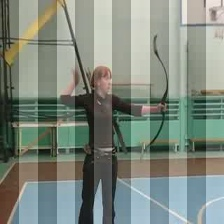}
         \caption{SIG}
         \label{fig:SIG}
     \end{subfigure}
     \hfill
     \begin{subfigure}[b]{0.12\textwidth}
         \centering
         \includegraphics[width=\textwidth]{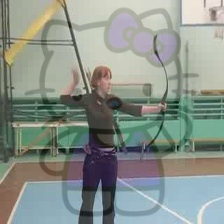}
         \caption{Blend}
         \label{fig:Blend}
     \end{subfigure}
     \hfill
     \begin{subfigure}[b]{0.12\textwidth}
         \centering
         \includegraphics[width=\textwidth]{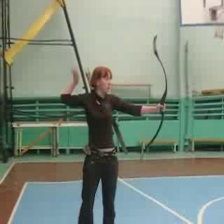}
         \caption{WaNet}
         \label{fig:WaNet}
     \end{subfigure}
     \hfill
     \begin{subfigure}[b]{0.12\textwidth}
         \centering
         \includegraphics[width=\textwidth]{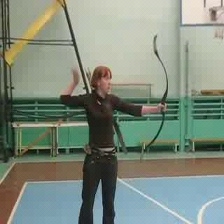}
         \caption{FT-trojan}
         \label{fig:FT-trojan}
     \end{subfigure}
\caption{Examples of Backdoor Triggers}
\label{fig:trigger_visualization}
\end{figure*}

All the backdoor triggers in this paper are visualized in Fig.~\ref{fig:trigger_visualization}.
We evaluate the imperceptibility of triggers to human perception using PSNR and SSIM, standard metrics in image quality assessment. 
For a more accurate evaluation, we employ localized quality metrics, with further details provided in Apdx.~\ref{sec:imperceptibility_more}.
Tab.~\ref{tab:imperceptibility} shows the results of all triggers.
Triggers from BadNet, Blend, SIG, and WaNet are relatively obvious to the human eye, while those from the proposed attack and FT-trojan are more imperceptible. 
The heightened imperceptibility arises 
since both attacks introduce triggers by perturbing the frequency domain, leading to minimal alterations per pixel. 
FTtrojan is slightly more imperceptible than ours, due to its gentler frequency domain perturbations. This also explains its relatively lower ASR.



\begin{table}[t] 
\centering
\small
\begin{tabular}{{p{.8cm}p{.6cm}p{.8cm}p{.8cm}p{.6cm}p{.8cm}p{1.1cm}}}
\toprule
\hline
Metric & DFT & BadNet & Blend & SIG & WaNet & FTtrojan \\
\hline
PSNR & \textbf{41.6} & 36.3 & 20.2 & 28.7 & 34.3 & \textbf{47.4} \\
SSIM & \textbf{0.972} & 0.173 & 0.435 & 0.515 & 0.842 & \textbf{0.982} \\
\hline
\bottomrule
\end{tabular}
\caption{Imperceptibility of all backdoor triggers measured by PSNR and SSIM.}
\label{tab:imperceptibility}
\end{table}

\subsection{Collateral Damage}
\begin{figure}
    \centering
    \includegraphics[width=.48\textwidth]{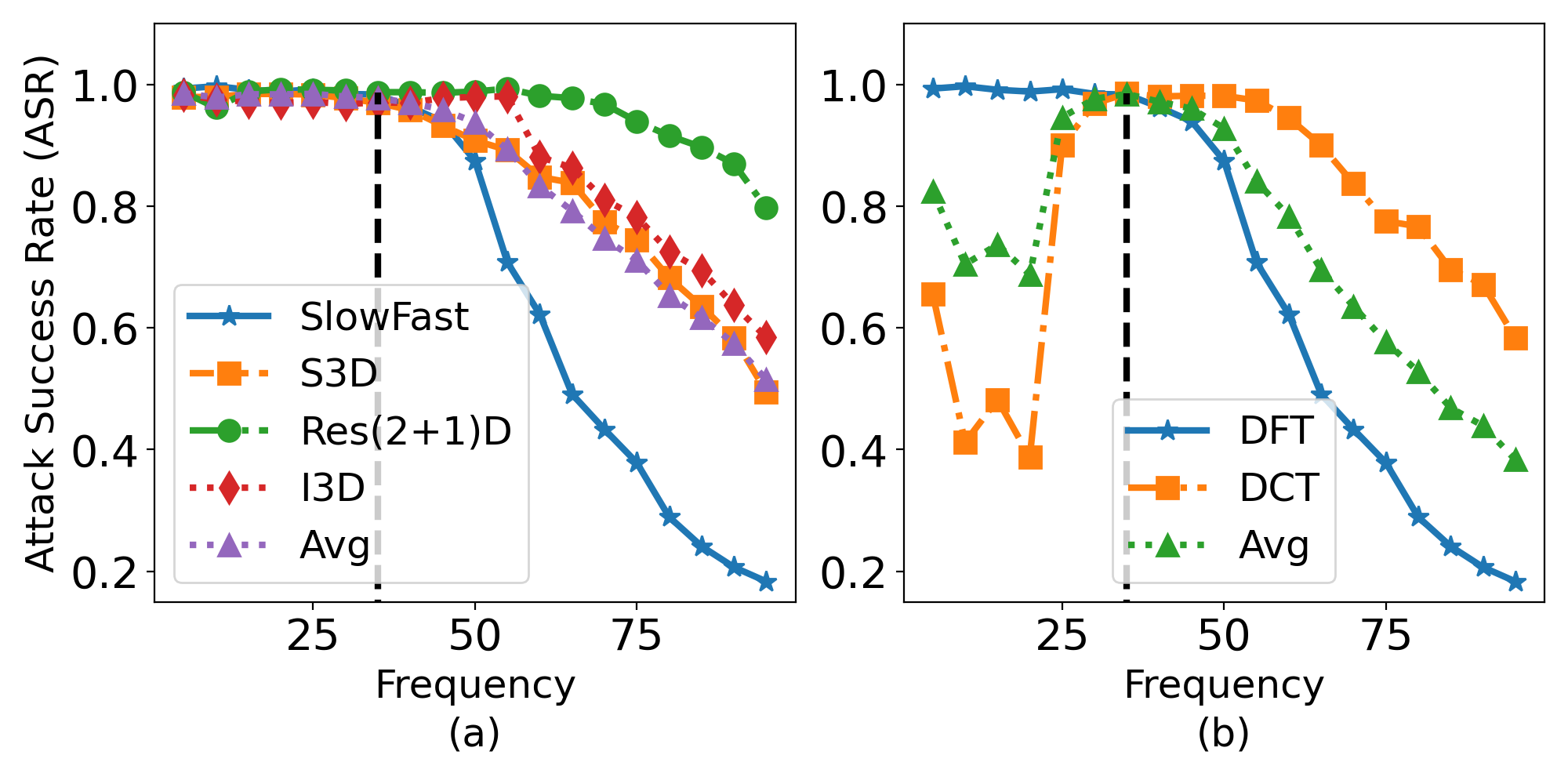}
    \caption{(a) Collateral damage on SlowFast, Res(2+1)D, S3D, and I3D trained on UCF-101 poisoned by the our DFT attack. (b) Collateral damage on SlowFast trained on UCF-101 poisoned by the proposed attack using DFT and DCT.}
    \label{fig:collateral_damage}
\end{figure}

Collateral damage refers to a phenomenon where perturbations in specific areas of the transformed domain could unintentionally activate the attack, even if they {\it mismatch} those intentionally introduced during training. We observe this phenomenon in our experiments and illustrate it by presenting the test ASR as a function of $k_0$ (in Fig.~\ref{fig:collateral_damage}), where the test instances are triggered on the set $\{k_0, \dots, k_0+9\}\times X\times Y$. 
The triggered instances are fed to four compromised models trained on UCF-101\footnote{Note that we only vary the frequencies for perturbation at test-time, and the compromised models are untouched.} and the results are shown in Fig.~\ref{fig:collateral_damage}(a).
The black dashed line denotes the frequencies perturbed during training ($k_0=35$).
The figure shows that the backdoor is successfully activated by lower-frequency perturbations for any model.
The ASR gradually declines as the perturbation affects higher frequencies.
Specifically, the ASR of SlowFast drops rapidly when the perturbed frequency exceeds 50, whereas the ASR of Res(2+1)D remains high.
This suggests that Res(2+1)D might be more vulnerable to adversarial perturbations compared to other model architectures, while SlowFast demonstrates relatively higher robustness.

We attribute this phenomenon to additional operations other than trigger embedding, such as pixel clipping.
These operations would introduce unintended perturbations to all components in the transformed representation.
Hence, during poisoned training, the victim DNN might inadvertently associate these unintended perturbations with the target class.
We further illustrate the results for attacks using DFT and DCT in Fig.~\ref{fig:collateral_damage}(b)
While both DFT and DCT exhibit this collateral damage, 
their effects manifest differently across frequency bands.
Specifically, DFT's unintended effects are primarily concentrated in the lower frequencies, whereas DCT shows these effects more in the mid-frequency range.

\subsection{Case Study: ASR v.s. Influential Factors}\label{sec:case study}

\begin{figure}
    \centering
    \includegraphics[width=.45\textwidth]{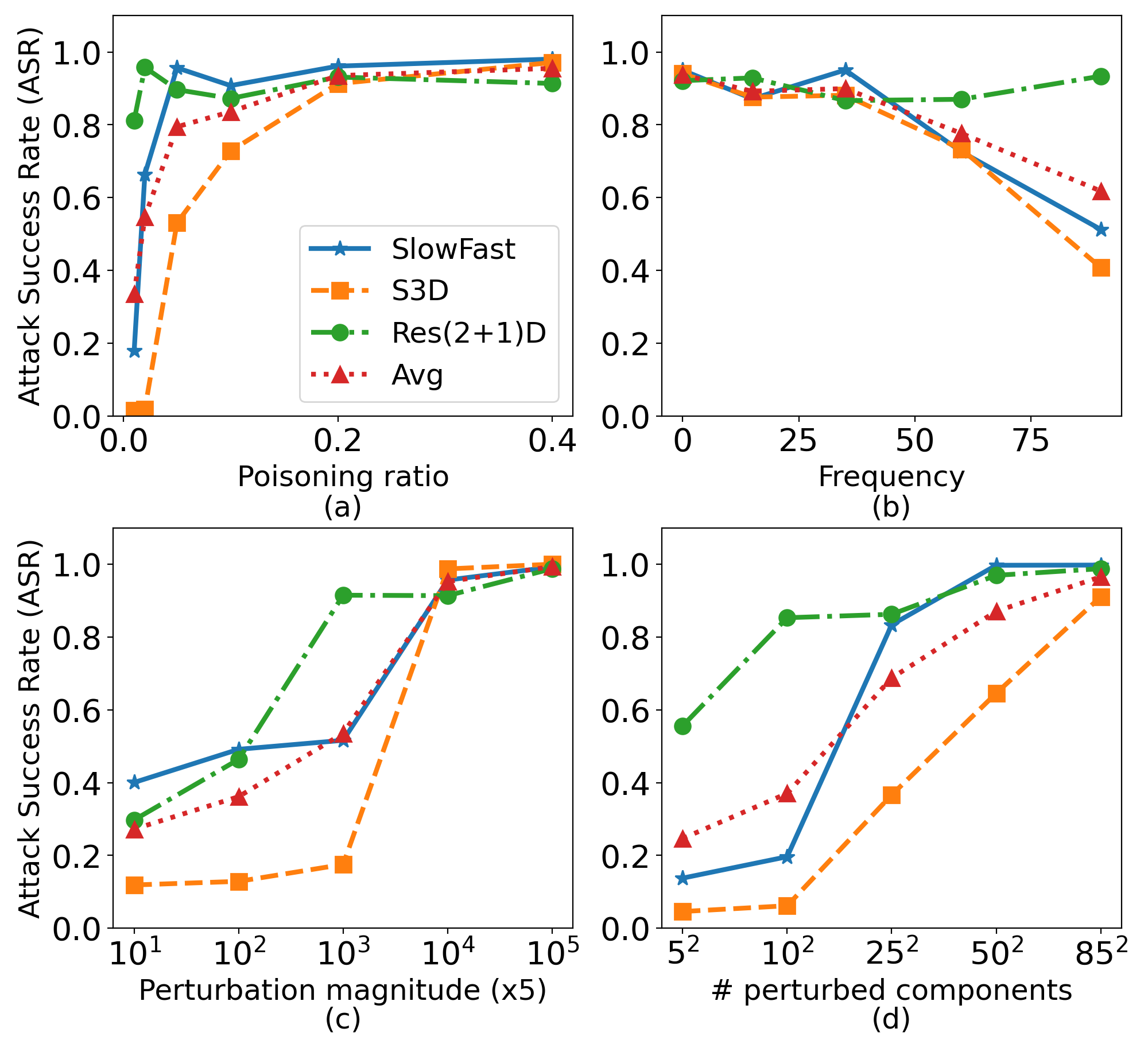}
    \caption{The ASR of various models trained on UCF-101 as a function of (a) poisoning ratio (b) frequencies for adding perturbation (c) perturbation magnitude (d) the number of perturbed components.}
    \label{fig:ASR_vs_4_factors}
\end{figure}

In this section, we examine the impact of various influential factors of the attack on the learning of backdoor mapping and model vulnerability.
Fig.~\ref{fig:ASR_vs_4_factors} illustrates how ASR is affected by four different factors. These experiments were conducted on the UCF-101 dataset compromised by a DFT-based attack, with more detailed settings available in Apdx.~\ref{sec:case_study_apdx}. The results suggest that \textit{attackers can easily choose suitable attack hyper-parameters for an effective attack across various models}, highlighting the importance of strengthening the defenses of video recognition systems.


\subsubsection{The Poisoning Ratio.}

The poisoning ratio represents the fraction of training samples under the manipulation of the attacker.
Unsurprisingly, the ASRs for all the models increase as the attack is strengthened. 
With just 5\% of the training data poisoned, both SlowFast and Res(2+1)D are compromised with ASRs greater than 90\%. 
Notably, even if the attacker merely manipulates 1\% of the training data, the attack is effective to Res(2+1)D with an ASR of around 80\%, while it is hard for the other models to build the backdoor mapping.
We believe Res(2+1)D is more susceptible to the adversarial attack compared with other CNN-based models. As a result, it prioritizes learning the backdoor mapping over the normal mapping during training, as shown in Fig.~\ref{fig:acc} in Apdx.~\ref{sec:case_study_apdx}.


\subsubsection{Frequency.}
The frequency refers to a range of frequencies $F = \{k_0, \dots, k_0+9\}$ where the attacker adds perturbations during \textit{training}, and we fix the length of the range at 10.
Fig.~\ref{fig:ASR_vs_4_factors} (b) displays ASR as a function of $f_s$.
Generally speaking, low-frequency components in images and videos represent the primary content, including large-scale structures, broad shapes, and general illumination. By contrast, high-frequency components capture details, edges, and textures.
It is not surprising that the attack becomes less effective as the perturbed frequency increases (except for Res(2+1)D), since perturbing low-frequency components generates more salient features than high-frequency components.
However, there is still a sufficiently large range of frequencies (0-50) for devising effective attacks.
Similar to the observation on the poisoning ratio, Res(2+1)D is vulnerable to a broader range of frequencies than the other models.


\subsubsection{Number of perturbed components.}
The total number of perturbed components over all selected frequencies is the size of $\mathcal{I}$ defined in Eq.~\ref{eq:trigger_embedding_function}.
These components are randomly chosen.
Similar to the observation on perturbation magnitude, the ASR monotonically increases with the number of components being altered.
With only 625 components (1.2\% of the total components in a $224\times224$ spectrum) in each frequency being perturbed, the attack successfully compromises the Res(2+1)D and SlowFast with ASR of around 80\%. 
S3D is resistant to the number of perturbed components -- the backdoor is planted with 7225 components being perturbed (14\% of the total components).

\subsubsection{Perturbation magnitude.}
The perturbation magnitude (denoted as $\delta$ in the trigger embedding function given by Eq.~\ref{eq:trigger_embedding_function}) represents the amount of change applied to each selected component.
The ASR monotonically increases with the perturbation magnitude: as the perturbation magnitude rises, the trigger becomes more salient in the original space.
Besides, the ASR gets a significant boost from a magnitude of $5\cdot10^{3}$ to $5\cdot10^{5}$. 
Hence, from the aspect of the attacker, choosing the right perturbation magnitude is straightforward -- with a few components being perturbed, applying a higher magnitude of perturbation can lead to more potent attacks.

\section{Limitation and Future Work}

In this paper, we focus on all-to-one attacks for the following reasons.
First, training video recognition models is inherently challenging given the increased complexity of video data compared to traditional image data.
Second, under attack settings such as many-to-one and one-to-one, there may not be a sufficient number of perturbed training samples to establish a solid mapping from the backdoor trigger(s) to the desired target classes.
However, many-to-one backdoor attacks would be more practical in scenarios, \textit{e.g.}, sign language translation.
The attacker would only aim to map a set of few words to another word(s) to introduce misinformation to the expression. 
Besides, there is no \textit{feasible} backdoor defense strategy proposed for complex triggers and complicated datasets. UNICORN~\citep{UNICORN} proposes a general framework to estimate a potential backdoor pattern embedded in any transformed space. 
However, it is practically infeasible due to the extremely high computation cost for estimating the trigger and the transform methods.
Also, it fails to detect if a given model is backdoor compromised.
We leave addressing the above problems as future works.

Finally, while the proposed attack exhibits collateral damage, it remains an open question: \textit{Does collateral damage has an impact on the stealthiness of the proposed attack under defenses or the robustness of victim models?} We conjecture that the answer is yes. Because WaNet~\citep{WaNet}, may also have collateral damage due to the clipping technique. This work indicates that without additional robust design such as noise mode, the attacked model is easy to be detected.


\if{0}
Besides, we observe that the proposed attack exhibits collateral damage, and theoretically analyze the underlying causes of this phenomenon. 
However, it remains an open question: \textit{Does collateral damage has an impact on the stealthiness of the proposed attack under defenses or the robustness of victim models?}
\citet{WaNet} proposed WaNet, a warping-based backdoor trigger. They also apply functions like clipping to eliminate invalid pixels brought by the warping effect.
They mentioned that, to avoid being caught by backdoor detection methods, they utilize clean poisoning to enhance stealthiness.
We conjecture this relates to the collateral damage -- pixel clipping causes unintended perturbations in the representation of the image in an unknown space, and such unintended perturbation is learned as part of the backdoor mapping by the victim classifier. 
Hence, the attack is detected as long as the unintended perturbation is estimated.
\fi

\section{Conclusion}
In this paper, we propose a general framework for embedding an imperceptible, temporal-distributed backdoor trigger in videos. 
Notably, it exhibits invisibility not only to human eyes, but also to current backdoor defense strategies.
Empirical experiments across various benchmark datasets and popular video recognition model architectures demonstrate the effectiveness of our attack. 
Furthermore, we explore the impact of several factors on the effectiveness of the proposed attack, providing an enriched perspective on the vulnerabilities in video recognition systems.

\bibliographystyle{apalike}
\bibliography{aaai24}

\appendix

\begin{center}
    {\Large \bfseries Temporal-Distributed Backdoor Attack Against Video Based Action Recognition: Appendix}
\end{center}

\section{Basis Construction for Additional Transforms}\label{sec:DWT}

In this section, we introduce the basis construction of other two methods, namely discrete wavelet transform, and random transform.
Recall that $\{\mathbf{e}_{n_0,n_1,n_2}\}_{n_{\iota}\in [N_{\iota}],\iota\in[3]}$ denote the standard basis of
single-channel video, i.e., $\mathbf{v} = \sum_{n_0,n_1,n_2} v(n_0,n_1,n_2)\mathbf{e}_{n_0,n_1,n_2}$.


\noindent{\bf Discrete Wavelet Transform (DWT).} DWT captures  localized information in video which is not realized by DFT or DCT. 
In this paper, we adopt Daubechies 1 (db1) wavelet transform. The basis $\mathbf{B}_{\text{W}} = \{ \mathbf{b}_{k_0,k_1,k_2}\}_{k_{\iota}\in[N_{\iota}],\iota\in[3]}$ of DWT is defined as follows. For any $k_{\iota}\in[N_{\iota}/2]$, $m_{\iota}\in\{0,1\}$, 

\begin{align*}
    &\mathbf{b}_{2k_0+ m_0 ,2k_1+m_1,2k_2+m_2} \\
    &\quad = \frac{1}{2\sqrt{2}}\sum_{\ell_0,\ell_1,\ell_2\in\{0,1\}}\mathbf{e}_{2k_0+\ell_0, 2k_1+ \ell_1, 2k_2+ \ell_2} \prod_{\iota=0}^2(-1)^{m_{\iota}\ell_{\iota}}.
\end{align*}

We note that the basis with $m_{\iota}=0$ represents the low-pass filter and the basis with $m_{\iota}=1$ represents the high-pass filter.

\noindent{\bf Random Transform (RT).} RT is a generalization of previous classic transforms. Specifically, we choose three random (invertible) matrices $\{\mathbf{M}^{\iota}\}_{\iota=0}^{2}$. The transform basis $\mathbf{B}_{\text{R}} = \{\mathbf{b}_{k_0,k_1,k_2}\}_{k_{\iota}\in[N_{\iota}],\iota\in[2]}$ is defined as follows.
\begin{align*}
    \mathbf{b}_{k_0,k_1,k_2} = \sum_{n_0,n_1,n_2} \mathbf{e}_{n_0,n_1,n_2}\prod_{\iota=0}^2\mathbf{M}^{\iota}(k_{\iota}, n_{\iota}),
\end{align*}
where $\mathbf{M}^{\iota}(k_{\iota}, n_{\iota})$ is the value at the $k_{\iota}$-th row and the $n_{\iota}$-th column of the matrix $\mathbf{M}^{\iota}$.

\if{0}
In this section, we provide the exact discrete wavelet transform (DWT) formulation for video data. 

For any $\iota\in[3]$, we assume $N_{\iota}$ is an even number. Then, for any $n_{\iota}\in[N_{\iota}/2]$, $k_{\iota}\in\{0,1\}$, the basis for DWT is defined as follows.

\begin{align*}
    &\mathbf{b}_{2n_0+ k_0 ,2n_1+k_1,2n_2+k_2} \\
    &\quad = \frac{1}{2\sqrt{2}}\sum_{\ell_0,\ell_1,\ell_2\in\{0,1\}}\mathbf{e}_{2n_0+\ell_0, 2n_1+ \ell_1, 2n_2+ \ell_2} \prod_{\iota=0}^2(-1)^{k_{\iota}\ell_{\iota}}.
\end{align*}
\fi

\section{Video Recognition Systems}
Over the years, researchers have formulated four distinct categories of video recognition models: 2D CNN + RNN, 3D-CNN, Two-stream CNN, and Transformer-based models. The 2D CNN + RNN approach is one of the earliest, wherein the 2D CNN is employed to extract features from individual frames, while the RNN is utilized to capture temporal dependencies between frames\cite{cl1,cl2,cl3,cl4,cl5}. Subsequently, 3D-CNNs emerged, processing both spatial and temporal dimensions concurrently, which allows for the recognition of motion patterns across consecutive frames\cite{c3d, p3d, i3d, can}. A significant leap in this domain was the introduction of the Two-Stream CNN model. This approach leverages both spatial frames and optical flow, enabling it to comprehensively discern appearance and motion information\cite{two1, two2, two3, two4}. Transformer-based models, inspired by the success in natural language processing, have begun to make an impact in this domain, offering self-attention mechanisms to weigh the importance of different frames\cite{timesformer, swin}.

\section{Datasets, Training Settings, and Attack Settings} \label{sec:datasets, training settings and attack settings}
\textbf{Datasets}:
We consider two benchmark datasets used in video action recognition, \textbf{UCF-101}~\citep{UCF} and \textbf{HMDB-51}~\citep{HMDB}, and a sign language recognition benchmark, Greek Sign Language (\textbf{GSL}) dataset~\citep{gsl}.
UCF-101 encompasses 13,320 video clips sorted into 101 distinct action categories. 
Similarly, HMDB-51 contains 7,000 video clips categorized into 51 classes of action. 
GSL incorporates 40,785 gloss instances across 310 unique glosses. To reduce computation cost, we form a subset of GSL by instances from 50 randomly selected classes. \\
\textbf{Training Settings}: 
We train all the 3D CNN-based models on all the datasets for 10 epochs, using the AdamW optimizer \cite{AdamW} with an initial learning rate of $0.0003$ .
Following the common training strategy in video recognition~\citep{BD_AR} and for reducing computation cost, we down-sample the videos into 32 frames. 
The Timesformer model (a transformer-based model in video recognition) \cite{timesformer} is trained for 5 epochs, with the SGD optimizer \cite{SGD} and an initial learning rate of $0.003$.
We down-sample the videos to 8 frames, as we initialize the Timesformer model parameters with those trained on a 8-frame video datasest.
Due to computation cost, the Timesformer model is only trained on UCF-101 poisoned by the proposed attacks using DFT and DCT.
\\
\textbf{Attack settings}: \\
In this paper, we consider \textit{all-to-one} attacks.
We choose \textbf{class 0} as the target class, and randomly select 20\% of the training samples per class for the attacker's manipulation.
For comparison, we embed classic triggers proposed for image data, including \textbf{BadNet}~\citep{BadNet}, \textbf{Blend}~\citep{Targeted-Backdoor}, \textbf{SIG}~\citep{SIG}, \textbf{WaNet}~\citep{WaNet}, \textbf{FTtrojan}~\citep{FTtrojan}, in each frame of the video.
We follow their poisoning pipeline and appropriately modify the attack hyper-parameters to achieve effective attacks.

For the \textbf{proposed attack}, we apply \textit{DFT}, \textit{DCT}, \textit{DWT}, and \textit{RT} for trigger generation.
\begin{enumerate}
    \item  For the attacks with \textit{DFT}, we select a subset $\mathcal{I} = \{35, 36, \dots, 44\}\times X\times Y\subset [N_0] \times [N_1]\times[N_2]$ with both $X,Y$ randomly selected and size $25$ in the frequency domain. We perturb the selected components with $\delta=50,000$.
    \item For \textit{DCT}, the set of component indices $\mathcal{I}$ is the same as DFT. We set the perturbation magnitude at $\delta=50$. 
    \item For \textit{DWT}, we employ the Daubechies 1 (db1) wavelet for transformation, which is defined in Sec.~\ref{sec:DWT}. The indices of the perturbed components are defined as $\mathcal{I}= \{12, \dots, 21\}\times[N_1] \times [N_2]$. We set the perturbation magnitude at $\delta=10$.
    \item For RT, the random matrices $\mathbf{M}^{\iota}$ are randomly generated square matrices and the values are between 0-1. We choose the perturbation magnitude of $\delta=30$, with $\mathcal{I}$ the same as DFT. 
\end{enumerate}
After inverting the altered representations, we create valid videos by taking the magnitude of complex numbers and clipping pixel intensities within $[256]$.

\begin{figure}[ht!]
\centering
     \begin{subfigure}[b]{0.15\textwidth}
     \centering
     \includegraphics[width=\textwidth]{figure/attack_examples/fft.png}
     \caption{DFT}
     \label{fig:DFT}
     \end{subfigure}
     \hfill
     \begin{subfigure}[b]{0.15\textwidth}
         \centering
         \includegraphics[width=\textwidth]{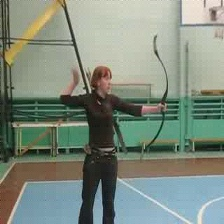}
         \caption{DCT}
         \label{fig:DCT}
     \end{subfigure}
     \hfill
     \begin{subfigure}[b]{0.15\textwidth}
         \centering
         \includegraphics[width=\textwidth]{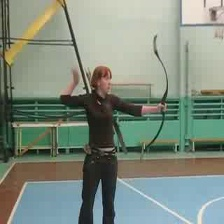}
         \caption{DWT}
         \label{fig:DWT}
     \end{subfigure}
     \hfill
     \begin{subfigure}[b]{0.15\textwidth}
         \centering
         \includegraphics[width=\textwidth]{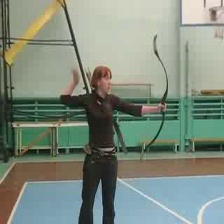}
         \caption{RT}
         \label{fig:RT}
     \end{subfigure}
\caption{Examples of backdoor triggers of the proposed attack using different transform methods.}
\label{fig:trigger_visualization_4_transforms}
\end{figure}

The hyper-parameter settings of the \textbf{baseline attacks} are as follows:
\begin{enumerate}
    \item For \textit{BadNet}, We embed a $21\times21$ white square in the top right corner of each frame.
    \item In the case of \textit{Blend},the Hello Kitty figure, as utilized by \citet{BD_AR}, is blended with each video frame using a blending coefficient of $\alpha=0.15$. That is, each frame $\mathbf{v}(t)$ in the triggered video becomes $\tilde{\mathbf{v}}(t) = (1-\alpha)\times\mathbf{v}(t) + \alpha \times \mathbf{p}$, with $\mathbf{p}$ representing the trigger.
    \item We embed the sinusoidal signal trigger of \textit{SIG} in each frame using the code released by \citet{NAD}.Here, the parameters are set as $\Delta=20$ and $f=6$.
    \item For WaNet, we set $k=8$ and $s=1$ for the warping field.
    \item For FTtrojan, we individually apply DCT on each frame, and add perturbation with magnitude of 30 to frequency 15 and 31 in each spectrum.
\end{enumerate}
All triggers are visualized in Fig.~\ref{fig:trigger_visualization} and \ref{fig:trigger_visualization_4_transforms}.
For all the attacks, the triggers are embedded into the down-sampled videos.


\section{Backdoor Detection Methods}\label{sec:detection settings}
we consider a \textit{pre-training} detection method, AC \cite{AC}. It detects and then removes outliers in the training set. For each class, it applies k-means to cluster the training samples into 2 clusters based on the poisoned DNN's internal layer activations. The smaller cluster of the two is deemed formed by poisoned samples if the silhouette score of the clustering is higher than a threshold. We set the detection threshold at 0.1, as suggested in their paper. Besides, we apply the detection on the \textit{penultimate layer activations} which are reduced to 10 components using Independent Component Analysis (ICA). To reduce computational cost, we only apply AC on class 0 -- the true target class.

For \textit{post-training} detection methods, we consider NC-D \cite{NC}, PT-RED \cite{PT-RED}, and TABOR \cite{TABOR}. They do not have access to the original training set, and detect the target class(es) as the one(s) corresponding to the outlier(s) among the reverse-engineered triggers.
The test statistic for anomaly detection is the modified z-score computed on pattern norms \cite{NC}.
We first compute the reciprocal of the norms of the estimated triggers -- $l_0$ norm for triggers estimated by NC-D and TABOR, and $l_2$ norm for triggers estimated by PT-RED.
Then we calculate the test statistic as the follows: (i) Compute the absolute deviation of all (reciprocal of) trigger norms from their median, and term the median of these absolute deviations as Median Absolute Deviation (MAD); (ii) Calculate the modified z-score for each sample by its absolute deviation from the median divided by MAD.
If the modified z-score is larger than 2 (a threshold suggested by \citet{NC}), we identify the corresponding class as the target class. 
Since these detection methods estimate a trigger for each class through optimization, these methods are extremely computationally expensive. Thus, we only reverse-engineer triggers and detect outliers for the first 10 classes of UCF-101, which contains the true target class 0.

We also consider a detection method deployed at \textit{test-time} -- STRIP \cite{STRIP}. It blends clean images to the input and observe the entropy of the posteriors: If the entropy is lower than a prescribed detection threshold, the input is deemed to be embedded with the trigger. Here, we set the detection threshold to achieve 15\% FPR on clean inputs, a choice which achieves a generally good trade-off between TPR and FPR.

\section{Backdoor Mitigation Methods}\label{sec:mitigation settings}
Several backdoor mitigation approaches are proposed to mitigate backdoor attacks at test time, without detecting if the given model is backdoor-compromised.
We apply a mitigation method, DBD \cite{DBD}, \textit{during DNN training}.
In the original framework, they decouple the poisoned training process by combining self-supervised learning, supervised learning, and semi-supervised fine-tuning. Specifically, they first train the backbone of a DNN on the unlabeled poisoned training set, then only train the fully connected layers on the labeled training set, finally fine-tune the whole DNN based on ``high-credible'' samples determined by the current model.
In our experiments, we simplified the framework to avoid the extensive costs of training on complex datasets repeatedly. We train the backbone of each video recognition model by predicting the rotation angle of rotated videos for 5 epochs and subsequently train the fully connected layers through supervised learning for another 5 epochs.
Both the learning rate and the optimizer used in these phases are the same as those specified in Sec.~\ref{sec:datasets, training settings and attack settings}.

Besides, we consider several \textit{post-training} mitigation methods.
We apply FP \cite{Fine-Pruning} on all the attacked models, which restores the mapping between the triggered instances and their true source classes by pruning neurons associated with the backdoor. To avoid significantly degrading the ACC, we prune 10\% of the neurons that are most likely related to the backdoor in layer 5 of SlowFast.
We also apply NAD \cite{NAD}, which preserves the classification function only for clean instances by knowledge distillation. The structures of teacher model and student model are the same as the poisoned model. The learning rate and optimizer used for obtaining the teacher model and knowledge distillation are the same as those stated in Apdx.~\ref{sec:datasets, training settings and attack settings}. For SlowFast, we do distillation on layer 2, 3, and 4.
NC-M unlearns the backdoor mapping by patching the DNN with correctly labeled triggered instances. We embed the triggers estimated by PT-RED into 30\% of the clean data possessed by the defender, and fine-tune the poisoned model on the correctly labeled dataset. The setting of fine-tuning is the same as the those stated in Apdx.~\ref{sec:datasets, training settings and attack settings}.

\section{Effectiveness of Attacks}\label{sec:effectivness_more}

\subsection{Using Different Transforms}
Apart from DFT considered in our main experiments (\textit{cf.} Tab.~\ref{tab:FT_ACC_ASR}), we also apply the proposed attacks using DCT and DWT against all the 3D CNN based video classifiers on UCF-101 and HMDB-51, with results shown in Tab.~\ref{tab:all_ASR}. 
For most of the cases, DCT and DWT based attacks achieve similar performance with the DFT based attack -- the attacks successfully compromises the victim models with ASRs above 95\%, and with ACCs close to the clean baselines.
However, there are some variations in attack effectiveness.
Specifically, both DCT and DWT based attacks cause a drop of around 10\% in ACC of the Res(2+1)D on HMDB-51, while the DFT based attack introduce negligible drop to the ACC. 
Conversely, on HMDB-51, the attack using DFT is less effective than those with DCT and DWT. With DFT, the ACC of I3D decreases by around 7\%, and the ASR is only 85\%. The ASRs of DCT and DWT based attacks are higher than 95\%, while the drop of ACCs is less than 2\%.

Due to computational constraints, we do not apply these attacks to the GSL dataset. 
Also, since RT is completely designed and implemented by ourselves, it lacks of the speed optimization applied in methods such as DFT, thus is time-consuming. Therefore, we only apply the RT based attack against SlowFast on UCF-101, yielding an ASR of 90.5\% and ACC of 78.7\%.


\begin{table*}[t] 
\small
\centering
\begin{tabular}
{p{.9cm}p{.9cm}p{.9cm}p{.9cm}p{.9cm}p{.9cm}p{.9cm}p{.9cm}p{.9cm}p{.9cm}}
\toprule
\hline
\multirow{2}{*}{Dataset} & \multirow{2}{*}{Attack} & \multicolumn{2}{c}{SlowFast} & \multicolumn{2}{c}{Res2+1D} &
\multicolumn{2}{c}{S3D} & \multicolumn{2}{c}{I3D} \\
\cline{3-10}
     & & ACC (\%) & ASR (\%) & ACC (\%) & ASR (\%) & ACC (\%) & ASR (\%) & ACC (\%) & ASR (\%) \\
\hline
\multirow{4}{*}{UCF101} & Clean & 84.5 & - & 77.4 & - & 90.6 & - & 89.0 & - \\
& DFT & 81.0 & 97.9 & 69.9 & 99.4 & 90.3 & 96.9 & 87.5 & 97.3 \\  
& DCT & 85.7 & 97.5 & 67.8 & 98.7 & 90.0 & 99.2 & 89.2 & 98.2 \\
& DWT & 83.0 & 99.7 & 67.6 & 99.0 & 90.6 & 99.9 & 85.7 & 99.9 \\
\hline
\multirow{4}{*}{HMDB51} & Clean & 60.6 & - & 53.6 & - & 69.3 & - & 66.6 & - \\
& DFT & 59.8 & 97.6 & 53.0 & 99.6 & 67.5 & 90.5 & 59.0 & 85.0 \\ 
& DCT & 59.6 & 97.1 & 43.3 & 96.5 & 69.3 & 97.7 & 65.7 & 96.8 \\
& DWT & 59.7 & 97.3 & 45.0 & 99.3 & 67.9 & 98.8 & 64.9 & 99.4 \\
\hline
\multirow{2}{*}{GSL} & Clean & 95.3 & - & 95.6 & - & 95.4 & - & 94.2 & - \\
& DFT       & 89.6 & 99.9 & 91.1 & 100.0 & 93.8 & 100.0 & 92.2 & 99.5\\
\hline
\bottomrule
\end{tabular}
\caption{ACCs and ASRs of SlowFast, Res2+1D, S3D, and I3d trained on UCF-101, HMDB-51, and GSL datasets poisoned by the proposed attack using DFT, DCT, and DWT.}
\label{tab:all_ASR}
\end{table*}

\subsection{Against the Transformer-based Model}
Compared with the 3D CNN based models, the attacks show similar performance against the TimeSformer model -- the compromised model yields a notable ASR on triggered instances while maintaining a high ACC.
Specifically, the TimeSformer trained on UCF-101 poisoned by the proposed attack using DFT achieves an ACC of 93.2\% and an ASR of 97.8\%.
The TimeSformer compromised by the DCT based attack yields an ACC of 92.2\% and ASR of 99.9\%.
These results demonstrate the efficacy of our attack on Transformer-based vision models.

\section{Trigger Visualization}

\begin{figure}[ht!]
\centering
    \begin{subfigure}[b]{0.15\textwidth} 
         \centering
         \includegraphics[width=\textwidth]{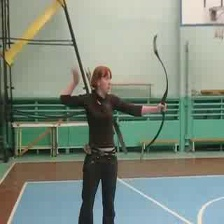}
         \caption{Clean}
     \end{subfigure}
     \hfill
     \begin{subfigure}[b]{0.15\textwidth} 
         \centering
         \includegraphics[width=\textwidth]{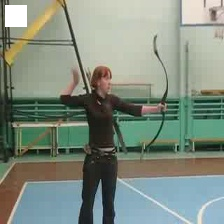}
         \caption{BadNet}
     \end{subfigure}
     \hfill
     \begin{subfigure}[b]{0.15\textwidth} 
         \centering
         \includegraphics[width=\textwidth]{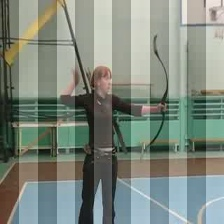}
         \caption{SIG}
     \end{subfigure}
     \hfill
     \begin{subfigure}[b]{0.15\textwidth}
         \centering
         \includegraphics[width=\textwidth]{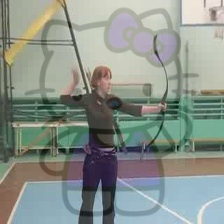}
         \caption{Blend}
     \end{subfigure}
     \begin{subfigure}[b]{0.15\textwidth}
         \centering
         \includegraphics[width=\textwidth]{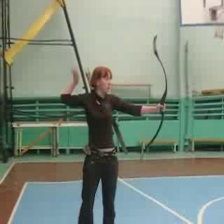}
         \caption{WaNet}
     \end{subfigure}
     \hfill
     \begin{subfigure}[b]{0.15\textwidth}
         \centering
         \includegraphics[width=\textwidth]{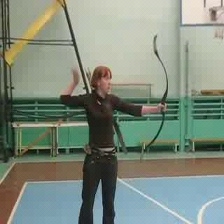}
         \caption{FT-trojan}
     \end{subfigure}
\caption{Examples of classic backdoor triggers.}
\end{figure}

\section{Trigger Imperceptibility}\label{sec:imperceptibility_more}

We measure the imperceptibility of a trigger to human observers by two common metrics for image quality degradation -- \textbf{PSNR} \cite{psnr_ssim} and \textbf{SSIM} \cite{ssim}.
\textit{However, these two metrics might not always align with human visual perception, especially when the types of perturbations are fundamentally different.}
For instance, BadNet creates a localized, overt pattern, but it yields higher PSNR and SSIM scores. The reason is that, the embedded white square is relatively small -- constituting approximately 0.9\% of the entire image (frame). 
As the bulk of the image (frame) remains unaltered, the values of both metrics are largely influenced by this untouched portion. 
Hence, we attempt to obtain a more accurate measure of pattern imperceptibility by \textit{localized quality metrics}. We calculate PSNR and SSIM over sliding windows in size of $28\times28$, and present the minimum, maximum, mean, and standard deviation (std) of these windows in Tab.~\ref{tab:imperceptibility_more}.
The metrics for each window are averaged across the whole test set. 

Although BadNet exhibits the highest averaged PSNR and SSIM values across the windows, its minimum value over the windows are relatively low, indicating the salience of the region with the trigger embedded.
Besides, its standard deviation is the largest among all the triggers, consistent with the trigger embedding strategy mentioned above.
As expected, SIG and Blend performs poorly in terms of both the minimum and average metrics, given they are globally obvious patterns.
WaNet also performs poorly in some local regions -- as the warping effect only affects certain portions of the frame.
By contrast, triggers generated by the proposed attack and FTtrojan demonstrate enhanced imperceptibility, with consistently high metrics across all windows. 
The heightened imperceptibility arises since both attacks introduce triggers by perturbing the frequency domain, leading to minimal alterations per pixel. 
FTtrojan is slightly more imperceptible than ours, due to its gentler frequency domain perturbations, which also explains its relatively lower ASR.

\begin{table*}[t] 
\centering
\begin{tabular}{llcccccc}
\toprule
\hline
\multicolumn{2}{c}{Metric} & DFT & BadNet & SIG & Blend & WaNet & FTtrojan \\
\hline
\multirow{4}{*}{PSNR} 
& min & 41.55 & 36.31 & 20.23 & 28.72 & 34.29 & 47.42 \\
& max & 46.51 & 98.13 & 24.51 & 30.58 & 45.08 & 48.63 \\
& mean & 42.87 & 97.16 & 22.24 & 29.17 & 38.82 & 48.03 \\
& std & 1.65 & 7.66 & 1.02 & 0.4775 & 2.369 & 0.3287 \\
\hline
\multirow{4}{*}{SSIM} 
& min & 0.9715 & 0.1727 & 0.4348 & 0.5145 & 0.8421 & 0.9824 \\
& max & 0.9882 & 1.0000 & 0.9076 & 0.8085 & 0.9809 & 0.9958 \\
& mean & 0.9834 & 0.9871 & 0.7257 & 0.715 & 0.9326 & 0.9907 \\
& std & 0.0045 & 0.1025 & 0.1396 & 0.0855 & 0.029 & 0.0036 \\
\hline
\bottomrule
\end{tabular}
\caption{Trigger Imperceptibility Measured by Localized PSNR and SSIM.}
\label{tab:imperceptibility_more}
\end{table*}

\begin{figure}
    \centering
    \includegraphics[width=.45\textwidth]{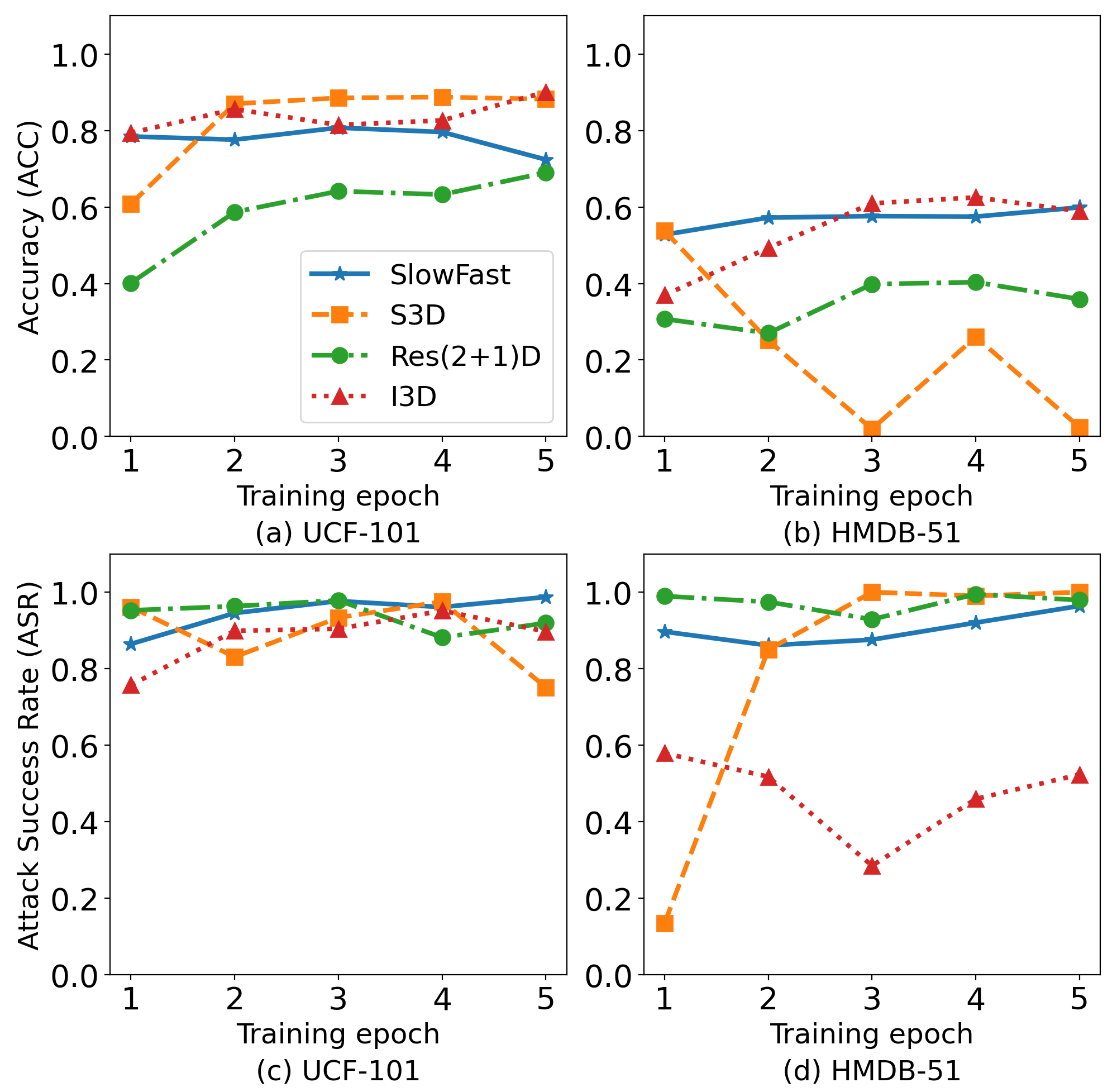}
    \caption{ACC and ASR of four 3D CNN-based models during the first 5 training epochs: (a) ACC on UCF-101, (b) ACC on HMDB-51, (c) ASR on UCF-101, (d) ASR on HMDB-51.}
    \label{fig:acc}
\end{figure}

\section{Experimental Setup for the Case Study}\label{sec:case_study_apdx}

In Sec.~\ref{sec:case study}, we examine the impact of various influential factors of the attack on the learning of backdoor mapping and model vulnerability. Fig.~\ref{fig:ASR_vs_4_factors} illustrates the ASR as a function of four factors: the poisoning ratio, the 
frequencies being perturbed, the perturbation magnitude, and the number of perturbed components.
In this section we elaborate the settings of the four sets of experiments. All experiments are conducted on the UCF-101 dataset poisoned by the proposed attack using DFT. 

\subsection{the Poisoning Ratio.}
To analyze the impact of the poisoning ratio on the learning of backdoor mapping,  we set the perturbed frequencies and components to $\mathcal{I} = \{35, 36, \dots, 44\}\times X \times Y$ with a fixed perturbation magnitude of $\delta=50,000$, where $X$ and $Y$ are the same as those specified in Sec.~\ref{sec:datasets, training settings and attack settings}. We choose the poisoning ratio in $\{0.01, 0.02, 0.05, 0.1, 0.2, 0.4\}$.

\subsection{Frequency.}
We experiment by adjusting the frequencies perturbed during training while maintaining a constant perturbation magnitude of $\delta=50,000$ and a poisoning ratio of 0.2. For each experiment, the set of indices of the perturbed frequencies and components is given by $\mathcal{I} = \{k_0, \dots, k_0+9\}\times X \times Y$, with $k_0$ chosen from $\{5, 15, 35, 60, 90\}$. $X$ and $Y$ are consistent with those specified in Sec.~\ref{sec:datasets, training settings and attack settings}.  

\subsection{Number of perturbed components.}
We experiment with varying sizes of $X$ and $Y$, which are randomly selected from the set in $\{5, 10, 25, 50, 85\}$. Different from the other experiments, the poisoning ratio if fixed at 0.05, since the learning of the backdoor mapping is slightly influenced by the number of perturbed components with a poisoning ratio at 0.2. We set the perturbation magnitude at $\delta=50,000$, and the perturbed frequencies at $\{35, 36, \dots, 44\}$. 

\subsection{Perturbation magnitude.}
To assess the influence of the perturbation magnitude $\delta$ on the attack effectiveness, we experiment with values from the set $\{50, 500, 5000, 50000, 500000\}$. We fix the poisoning ratio at 0.2, and the perturbed frequencies and components to $\mathcal{I} = \{35, 36, \dots, 44\}\times X \times Y$. $X$ and $Y$ remain the same as outlined in Sec.~\ref{sec:datasets, training settings and attack settings}.

\end{document}